\documentclass[10pt,twocolumn,letterpaper]{article}

\usepackage[pagenumbers]{cvpr} 
\usepackage{times}
\usepackage{epsfig}
\usepackage{graphicx}
\usepackage{amsmath}
\usepackage{amssymb}
\usepackage{booktabs}
\usepackage{mathrsfs}
\usepackage{bm}

\usepackage{algorithm, algorithmic}
\usepackage{mathdots}
\usepackage{multirow}
\usepackage{microtype} 
\usepackage{array}
\usepackage{setspace}
\usepackage{threeparttable}

\usepackage{graphicx}
\usepackage{diagbox}
\usepackage{wrapfig}
\usepackage{float}
\usepackage{bbm}
\usepackage{url}

\usepackage[dvipsnames]{xcolor}

\definecolor{cvprblue}{rgb}{0.21,0.49,0.74}
\usepackage[pagebackref,breaklinks,colorlinks,citecolor=cvprblue]{hyperref}

\usepackage{pifont}
\usepackage{makecell}

\definecolor{forestgreen}{RGB}{47, 159, 87}
\usepackage{pifont}
\usepackage{makecell}
\newlength\savewidth

\definecolor{gray}{rgb}{0.5, 0.5, 0.5}

\title{CustomListener: Text-guided Responsive Interaction for \\ User-friendly Listening Head Generation}

\author{Xi Liu\thanks{Equal contribution} ,  Ying Guo\textsuperscript{*},  Cheng Zhen, Tong Li, Yingying Ao, Pengfei Yan\thanks{Corresponding Author}  \vspace{2mm}\\
Vision AI Department, Meituan\\
\small \url{https://customlistener.github.io/}
}

\begin{document}
\maketitle
\begin{abstract}
Listening head generation aims to synthesize a non-verbal responsive listener head by modeling the correlation between the speaker and the listener in dynamic conversion.
The applications of listener agent generation in virtual interaction have promoted many works achieving diverse and fine-grained motion generation. However, they can only manipulate motions through simple emotional labels, but cannot freely control the listener's motions.
Since listener agents should have human-like attributes (e.g. identity, personality) which can be freely customized by users, this limits their realism. In this paper, we propose a user-friendly framework called CustomListener to realize the free-form text prior guided listener generation. To achieve speaker-listener coordination, we design a Static to Dynamic Portrait module (SDP), which interacts with speaker information to transform static text into dynamic portrait token with completion rhythm and amplitude information. To achieve coherence between segments, we design a Past Guided Generation module (PGG) to maintain the consistency of customized listener attributes through the motion prior, and utilize a diffusion-based structure conditioned on the portrait token and the motion prior to realize the controllable generation. To train and evaluate our model, we have constructed two text-annotated listening head datasets based on ViCo and RealTalk, which provide text-video paired labels. Extensive experiments have verified the effectiveness of our model.

\end{abstract}    
\section{Introduction}
\label{sec:intro}

Listening Head Generation (LHG) is to generate listener motions (face expressions, head movements, etc.) that respond synchronously to the speaker during face-to-face communication.
Good communication is a two-way process, and the roles of speakers and listeners are equally important \cite{cassell1999power11}.
Whether the listener responds promptly will affect whether the speaker can actively express the follow-up content \cite{kendon1970movement41,lafrance1979nonverbal43}.
Unlike speaker head generation (SHG) \cite{chen2019hierarchical10,ji2021audio21,richard2021meshtalk35,vougioukas2020realistic53}, listeners can only convey information through non-verbal feedback (nodding, frowning, etc.), and their responses are affected by both sides of the interaction (speaker's motion, tone, and listener's own personality and emotion state).
Currently, many applications, such as digital avatar generation \cite{chen2021high,li2021ai} and human-computer interaction \cite{zhang2021flow}, use the listener agent to simulate more realistic conversations, which has promoted many studies in LHG.

\begin{figure}[t]
   \includegraphics[width=\linewidth]{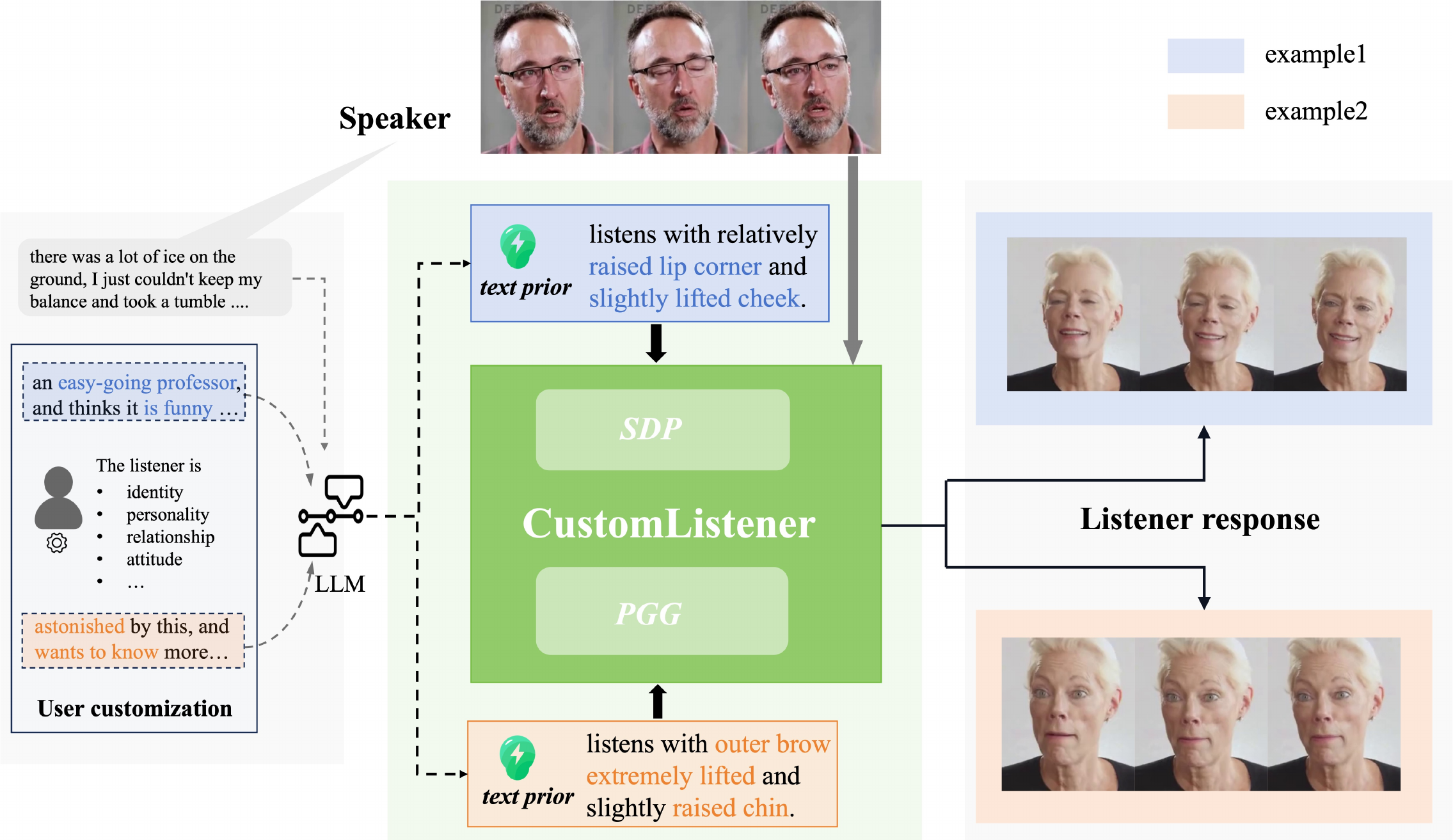}
   \caption{The process of text-guided listener generation in our CustomListener. The text prior provides the basic portrait style of the listener, which is input into CustomListener and combined with the speaker's information to obtain the listener's motions.}
   \vspace{-0.5cm}
   \label{fig:figure_1}
\end{figure}

Some early works manually integrate motions based on rules \cite{sonlu2021conversational,bohus2010facilitating,cassell1994animated}, or rely on 2D facial keypoints \cite{feng2017learn2smile,nojavanasghari2018interactive} to manipulate the head, but they are limited in the variety of motions and the detailed control ability.
Recent works \cite{vico,pchg,mfr,ELP,l2l} have introduced 3D face coefficients \cite{deep3d}, combined with the post-rendering module to achieve more precise control.
Specifically, RLHG \cite{vico} uses a sequence-sequence structure as the baseline method for decoding 3DMM coefficients. PCH \cite{pchg} further improves the render module to inpaint image artifacts.
L2L \cite{l2l} quantizes listener motions into the discrete one-dimensional codebook by VQ-VAE \cite{vqvae}, and ELP \cite{ELP} expands the codebook to a composition of several
discrete codewords to achieve more fine-grained control.
MFR-Net \cite{mfr} designs a feature aggregation module to achieve better identity preservation and motion diversity.

Although the above methods can combine speaker-listener information to generate diverse and fine-grained listener motions, they have restrictions in the explicit control of the listener agent's response.
Specifically, L2L \cite{l2l} focuses on the motion diversity but cannot control the motion under specific conditions.
RLHG \cite{vico}, PCH \cite{pchg}, MFR-Net \cite{mfr}, and ELP \cite{ELP} achieves controllable listener motion conditioned on the input attitude. However, the attitude is limited to a few fixed labels (eg. positive, neutral, negative), posing limitations on the generation of realistic listeners: 
First, simple labels are not enough to give accurate responses. For example, for negative attitudes, the two emotions of doubt and anger correspond to different facial expressions. 
More importantly, a realistic listener agent should have its own identity (parent, friend, teacher, etc.), personality (lively, calm, etc.) and behavioral habits (e.g. frowning when thinking), which users can customize and set in advance. However, simple labels cannot achieve the control under these conditions.

Considering the above limitations of simple labels, in this paper, we propose a user-friendly framework called CustomListener to enable the text-guided generation, as shown in Figure \ref{fig:figure_1}.
Users can pre-customize detailed attributes of the listener agent.
Then using a large language model, we combine speech content and user-customized attributes to obtain the text prior of the listener's basic portrait.
In CustomListener, we seamlessly incorporate speaker information while being guided by the text prior, so as to generate realistic listener responses that are controllable and interactive.

For the guidance of text prior, due to the interactive nature of communication, listener motions cannot simply be regarded as text-conditioned motion generation, but faces the following challenges: 1) \textbf{Speaker-listener coordination}: the text provides the basic portrait style of the listener's response. While to precisely express the listener's emotional empathy towards the speaker, the response not only needs to complete the text-specified motion, but also needs to adjust its completion timing and rhythm, and fluctuate with the speaker's semantics, intonation and movement amplitude.
2) \textbf{Listener motion coherence}: When motions conditioned on different text priors are combined into a long video, different segments should be coherent.
Due to the customization, apart from ensuring smooth motion switching, we also need to guarantee that the user-customized listener's behavioral habits reflected in past segments are maintained in the current segment. Maintaining the style of the past while still showing the motion of the current text can be challenging.

To solve the above challenges, in our CustomListener, we design the Static to Dynamic Portrait module (SDP) and the Past Guided Generation module (PGG) to achieve the \emph{coordination} and the \emph{coherence} respectively.
In SDP, we transform the static portrait provided by the text prior to be dynamic one, which interacts with audio to obtain the time-dependent information about the completion time and rhythm of the gradual motion changes, and refines the listener's motions to make the speaker-listener fluctuations relevant. In PGG, we generate a motion prior based on the similarity of dynamic portrait tokens between adjacent segments, which provides reference information for features that need to be maintained in the current compared to the past.
Then, the motion prior and the dynamic portrait token are input as conditions into the diffusion-based structure \cite{ddpm} with fixed noise at the segment connection, finally generating listener coefficients that enable coordination and coherence in interactions.

In summary, this paper has the following contributions:
\begin{itemize}
\setlength{\itemsep}{0pt}
\item We argue that simple labels are not sufficient to achieve freely controllable listener generation, and thus propose the CustomListener framework that can freely control listener head motions guided by the text prior.
\item 
We design a SDP module to achieve the speaker-listener coordination, so that the completion timing and rhythm of text-specified motions correspond to the speaker, and fluctuate with the speaker's semantics, tone, and movements.
\item For long-term generation, we design a PGG module to meet listener motion coherence between different text-prior-conditioned segments, so that the consistency of the listener's customized behavioral habits between segments can be maintained and the motion switching is smooth.
\item Extensive experiments on ViCo \cite{vico} and Realtalk \cite{realtalk} datasets confirm that our method can realize the state-of-the-art performance, and achieve the controllable and interactive realistic listener motion generation.
\end{itemize}
\section{Related work}
\begin{figure*}[t]
   \includegraphics[width=\linewidth]{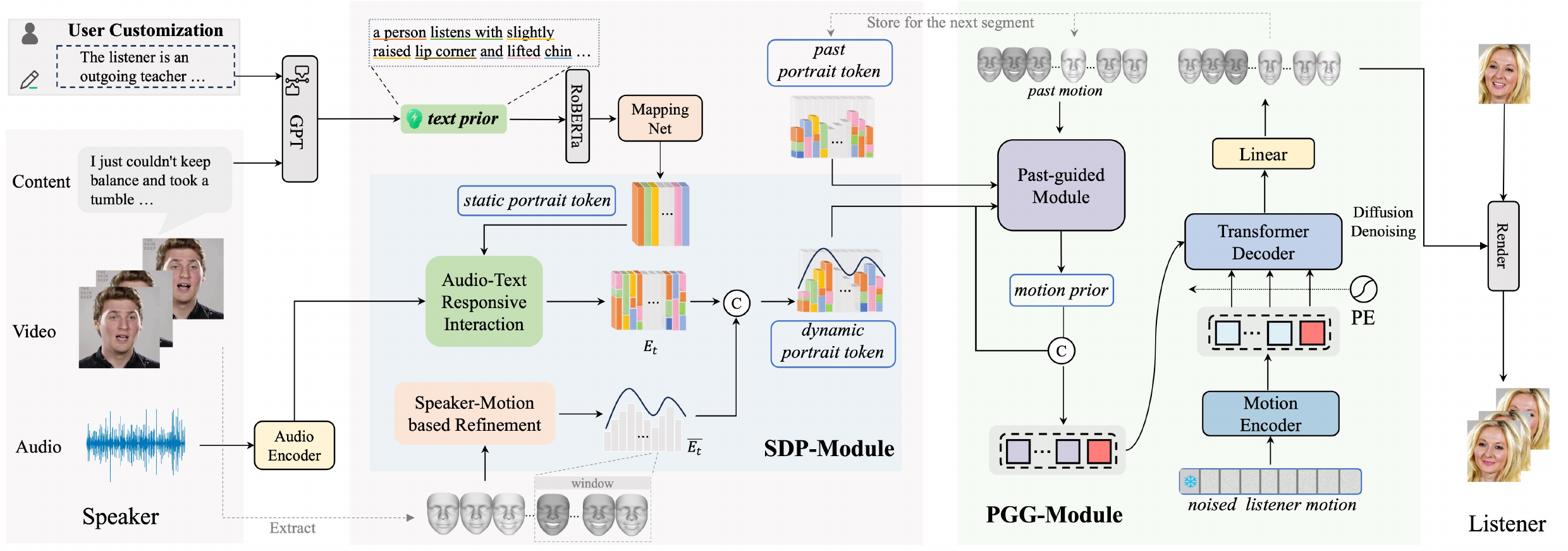}
   \caption{Overall framework of CustomListener. Given the text prior providing the listener's static portrait style, SDP-Module transforms the static portrait into a dynamic one. Then in PGG-Module, the dynamic portrait token are combined with motion prior generated from Past-guided Module and are utilized as conditions of the diffusion-based structure to realize the controllable generation. The 'C' in the figure denotes concatenation and the small pink squares denotes diffusion time-step token.}
   \vspace{-0.5cm}
   \label{fig:main}
\end{figure*}

\paragraph{Listening head generation (LHG):}Some early works \cite{sonlu2021conversational,bohus2010facilitating,cassell1994animated} use rule-based methods to manually incorporate interactive motions.
Later, data-driven methods \cite{feng2017learn2smile,nojavanasghari2018interactive} generate 2D motions based on facial keypoints or movement frequencies, but 2D information cannot fully represent the details of facial movements.

Recently, motion generation based on 3D face coefficients, coupled with the post-processing renderer, has gradually become mainstream due to its stronger representation ability of facial movements.
RLHG \cite{vico} proposes a representative ViCo dataset and constructs a baseline coefficient decoding method using a sequence-sequence structure.
After that, PCH \cite{pchg} assembles an enhanced renderer to make the generated results visually better and more stable by inpainting the image boundary and background.
Learning2Listen (L2L) \cite{l2l} proposes to use a sequence-encoding VQ-VAE \cite{vqvae} to learn the discrete codebook of listener motion.
ELP \cite{ELP} further utilizes a composition of multiple discrete motion-codewords to represent facial motion in a more fine-grained manner.
MFR-Net \cite{mfr} achieves the preservation of listener's identity and the diversity of generation.

Regarding the listener's motion status, L2L \cite{l2l} focuses on the diversity but cannot explicitly control specific actions.
RLHG \cite{vico}, PCH \cite{pchg}, MFR-Net \cite{mfr} achieve controllable generation conditioned on the input emotion, but the emotion is limited to a few labels and cannot specifically control the expression and posture of each emotion.
Although ELP \cite{ELP} represents the latent space under different emotions in a fine-grained manner through the codebook, what the user can control is still the input emotion rather than the fine-grained motions. In this way, the listener's response to each emotion may depend on the data distribution of the training set.

\vspace{-0.5cm}
\paragraph{Conditional motion generation:}The motion generation task derives applications under a variety of conditions. For example, body motion conditioned on text descriptions \cite{kim2023flame,petrovich2021action_actionlabel,yang2023synthesizing_actionlabel,petrovich2022temos,ghosh2021synthesis}, motion trajectory conditioned on scene image \cite{cao2020longscene}, and dance motion conditioned on music \cite{yang2023longdancediff}. Recently, large language models promote the development of free-form text conditioned generation based on auto-encoder \cite{petrovich2022temos}, hierarchical model \cite{ghosh2021synthesis}, and diffusion models \cite{kim2023flame}.
We also generate motions conditioned on free-form text.
However, for our responsive interaction, in addition to focusing on the matching of motion and text like the above methods, we also need to make the motion and the speaker's action, speech, tone and rhythm harmonious to express the listener's empathy. 
In our paper, we adopt the SDP module to achieve this harmony.
For long-term generation, to ensure the coherent fluency between clips under different text commands, some methods have adopted transition sampling \cite{synthesizing}, specifying noise \cite{diffusion_priors}, etc.
However, the coherence for the custom listener, which has some motion habits, is more complex.
It is not just a simple motion switch, but is related to past habits and styles, and our PGG module solves this challenge.

\section{Method}
\label{sec:method}

\subsection{Overview}
In our CustomListener framework, under the guidance of the fine-grained text prior, we can synthesize natural and controllable listener motions, which can seamlessly integrate the speaker's information (motion, speech content, audio) to give interactive non-verbal responses.
Let $L_{m}\in \mathbb{R}^{T\times 70}$ denote the listener motions to be generated, and $T$ is the number of frames, 70 denotes 64-dim expression and 6-dim pose.
Given speaker's motions $S_{m}\in \mathbb{R}^{T\times 70}$, audio $S_{a}$, speech content $S_{c}$ and text prior $R$, our framework can be formulated as:
\vspace{-0.2cm}
\begin{equation}
    L_{m} = Custom(S_{m},S_{a},S_{c};R) 
     \vspace{-0.2cm}
\end{equation}

The overview of our proposed framework is shown in Figure \ref{fig:main}.
Before generation, for each video segment, we use GPT to generate a text prior based on user-customized listener attributes and the speaker's speech content, which provides the listener's static portrait reference.
Then, we extract the speaker's audio features from the Audio Encoder and speaker's motion coefficients from a pre-trained face reconstruction model \cite{deep3d}.
Following this, we generate realistic listener motions through a two-stage process: (1) Static to Dynamic Portrait generation (SDP-Module): it can transform static portrait tokens into the dynamic through audio-text responsive interaction and speaker-motion based refinement. The dynamic portrait token includes the time-dependent information about the motion completion and rhythm, and makes the speaker-listener motion fluctuations correlated; (2) Past-guided Listener Motion Generation (PGG-Module): to produce coherent motions and maintain the listener's previous behavioral habits in long-term generation, we use a past guided module to produce the motion prior, combined with the diffusion denoising process with a fixed initial noise to generate the listener coefficients.
Finally, listener videos are obtained through the post-rendering process.

\subsection{Static to Dynamic Portrait generation}
\label{subsec:3.2}
In this section, we first introduce the generation of the static portrait-token, and then present SDP-Module, which can convert static tokens into dynamic one by two stages: (1) Audio-text Responsive Interaction: static portrait tokens are converted into time-dependent portrait tokens, which reflects the completion time and rhythm of motions according to the semantics and intonation in the audio, ensuring gradual motion changes; (2) Speaker Motion-based Refinement: it refines the listener's motions according to the fluctuations of speaker's motions, producing the final dynamic portrait token combined with the time-dependent portrait token.

\vspace{-0.4cm}
\paragraph{Static Portrait-token Generation}
To establish a better connection between the user-customized listener attributes (identity, personality, relationship, attitude, etc.) and the listener's motions, we leverage the powerful GPT for semantic understanding of both user-customized contents and speaker's speech contents, and generate the text prior that accurately describes the listener's expressions and poses.
Following this, we use RoBERTa \cite{roberta}, a robust and efficient model with high natural language understanding capability, to encode the text prior into text embeddings. To obtain portrait-related tokens, we further employ a mapping net to convert text embeddings into portrait tokens. We consider 
these portrait tokens to be static portrait tokens since the text prior describes the listener's static portrait (eg. expressions and poses) in the entire video clip.

\vspace{-0.4cm}
\paragraph{Audio-text Responsive Interaction} 
Since static portrait tokens from text prior provide a comprehensive representation of listener head in a video segment, it lacks the time-dependent information. Thus, the model would be limited to generating constant motion and cannot effectively convey gradual variation in movement without a specific design. To address this critical issue, we make use of the semantic knowledge in speaker's utterances, which serves as a guidance for motion shifts. Specifically, we choose the speaker audio to provide semantics instead of speech contents. The underlying intuition is the former contains not only semantic information but also the temporal information. This entire process can be split into two parts, part A and part B.
\begin{figure}[t]
     \centering
     \includegraphics[width=\linewidth]{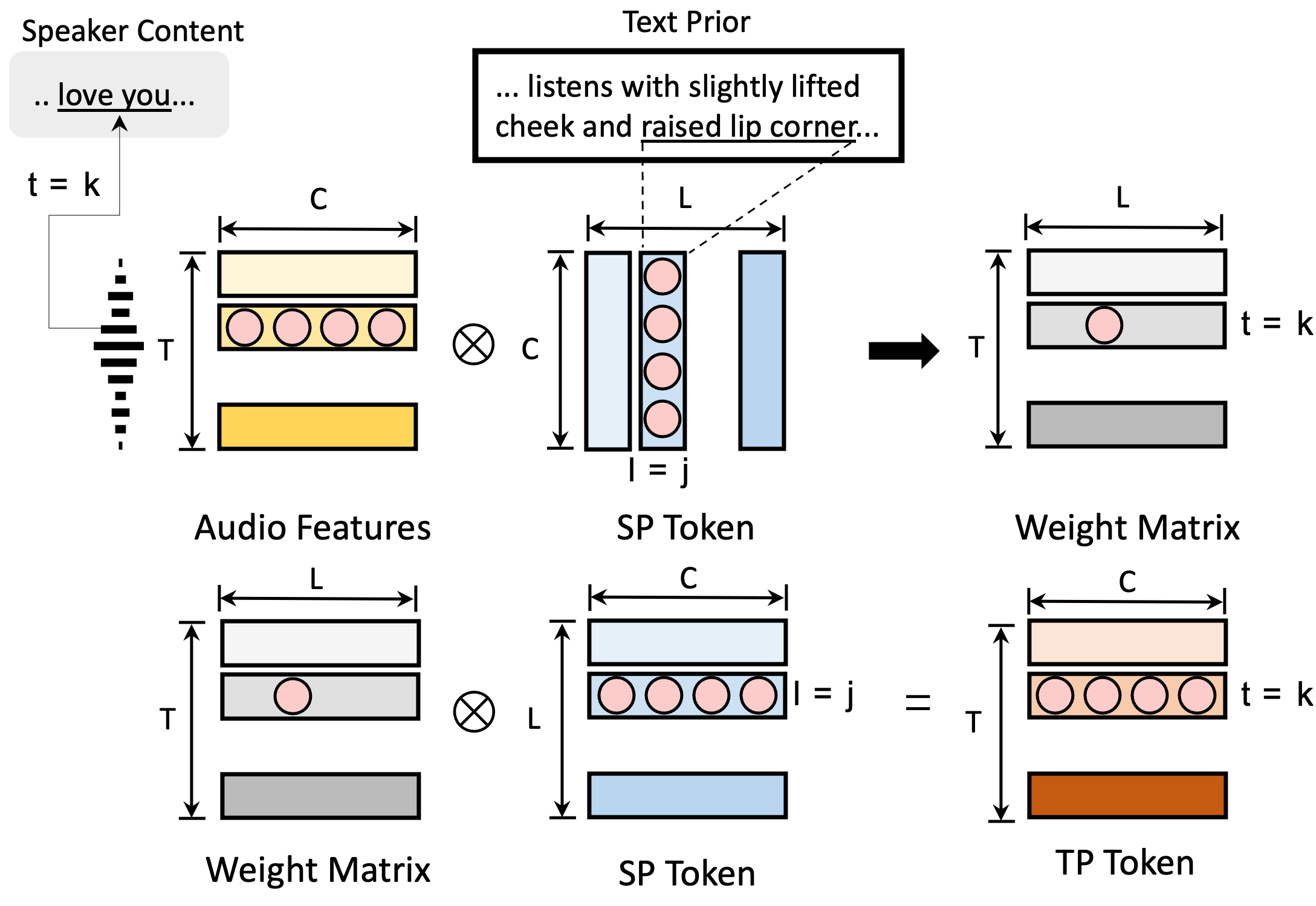}
     \caption{Illustration of Audio-text Responsive Interaction. We first generate a weight matrix by responsive interaction between audio features and static portrait-token (SP Token), and then generate time-dependent portrait token (TP Token) guided by weight matrix.}
     \label{fig:fig3}
    \vspace{-0.5cm}
\end{figure}

\textbf{Part A} In this part, we generate a weight matrix through responsive interaction between speaker audio features and the static portrait tokens obtained from text prior.
It provides the correlation degree between the speaker's semantics at each timestamp and the motion description in the text prior.
For example, as shown in Figure \ref{fig:fig3}, if the phrase ``love you" in the speaker's audio at time stamp $k$ exhibits the strongest correlation with the phrase ``raised lip corner" in the description texts, the corresponding weight value will reach its peak (indicated by a pink circle dot in the weight map). This weight matrix $M(A,H)$ can be formulated as Equation \ref{eq:softmax}.

\textbf{Part B} After obtaining a weight matrix $M(A,H)$, we can generate time-dependent portrait tokens based on static portrait tokens with the guidance of $M(A,H)$. Specifically, as illustrated in Figure \ref{fig:fig3}, the time-dependent portrait tokens (TP Token) at the time stamp $k$ is generated with the highest weighting assigned to the static portrait tokens (SP Token) at $l=j$, which corresponds to the phrase ``raised lip corner". This part can be formulated as Equation \ref{eq:weight}.
\begin{equation}
    \label{eq:softmax}
   M(A,H) = Sm(\frac{1}{\sqrt{\alpha }} (Linear(A)\times Linear(H)^{T })),
\end{equation}
\begin{equation}
    \label{eq:weight}
    E_{t}=\sum_{i=0}^{L-1} M(A,H)_{i,t} \cdot H_{i, \operatorname{j}=0: C-1}, t\in\left \{ 0,...,T-1 \right \}
\end{equation}
where $Sm$ denotes Softmax, $A\in \mathbb{R}^{T\times C}$ is the audio feature, $H\in \mathbb{R}^{L\times C}$ is the static portrait token, $M\in \mathbb{R}^{T\times L}$ is the weight map, $\alpha$ is a scaling factor, $E_{t}\in \mathbb{R}^{1\times C}$ is the time-dependent portrait token at time stamp $t$.
$L$ is the length of text embeddings, and $C$ is the feature dimension.

With the help of two parts above, we can obtain portrait tokens that contain time-dependent information about the rhythm of motion completion. This enables the model to learn the progressive changes in facial/head motions within a single video segment, as opposed to the constant motions.   

\vspace{-0.3cm}
\paragraph{Speaker Motion-based Refinement}
In human-to-human conversations, we believe that the amplitude of the speaker's motion fluctuation also has an impact on the changes in the listener's motion response\cite{richardson2008synchrony,mfr}.
Therefore, we utilize the past speaker's motion changes to refine the listener's motion amplitude.
This process can be well-designed formulated as:
\begin{equation}
   \overline{E_{t}} =\sum_{i=t-1-w}^{t-1}\exp \left|\frac{\partial S_{i} }{\partial i} \right|,
\end{equation}
\begin{equation}
   \widetilde{E_{t}} = E_{t} \oplus \overline{E_{t}},
\end{equation}
where $\oplus$ means concatenation, $E_{t},\overline{E_{t}},\widetilde{E_{t}}$ is time-dependent portrait token, speaker motion-weighted portrait token and dynamic portrait token at frame $t$, respectively. $S_{i}$ denotes speaker motion at frame $i$, $w$ is the time-window length. For efficiency, we set $w$ to 5, which means we only take the past five-frame speaker motions into account. For each time stamp $t$, the listener's portrait token $E_{t}$ is affected by the cumulative gradients of several previous speaker motions. This process can be illustrated in the lower part of the SDP-Module in Figure \ref{fig:main}. Notably, when generating portrait token at frame $t$, we only focus on the speaker motions up to frame $t-1$. Due to the listener and the speaker communicating concurrently, the listener's state at frame $t$ should not influenced by the speaker's motion at the same frame $t$.

\subsection{Past-guided Listenr Motion Generation}
The PGG-Module is to ensure that motions between segments are coherent, and the customized listener attributes in the previous segment can be maintained in the current segment.
It contains Past-guided Module and diffusion-based motion generation module, which can generate motion prior based on past motions and then synthesize listener motions. Following this, a renderer is used to obtain listener videos.

\vspace{-0.3cm}
\paragraph{Past-guided Module}
To produce long video with arbitrary length, we opt for a segment-by-segment generation approach during inference. However,  we have observed that there are unsmooth pose transitions between segments. This issue can be attributed to the neglect of correlations between adjacent segments. To address this, we take past motions into consideration. 

Since our final goal is to synthesize a customized listener, in addition to ensuring smooth transition, we also need to take the listener's past behavioral habits into account (e.g. frowning when thinking). Hence, we sought assistance from dynamic portrait tokens of two adjacent segments to guide the generation of motion prior. Specifically, the weight we assign to past motions at different times is determined by the similarity between portrait tokens in the current segment and those in the last segment. Formally, we calculate the motion weight and obtain the motion prior $G_{t:t+L}$ by:
\begin{equation}
   Sim_{t:t+L}=  E_{t:t+L} \times E_{t-L:t}^{T},
\end{equation}
\begin{equation}
   G_{t:t+L} = Sim_{t:t+L} \times P_{t-L:t},
\end{equation}
where $\times$ denotes matrix multiplication, $L$ is the segment length, $t$ is the time stamp, $Sim_{t:t+L}$ denotes the similarity between portrait tokens in the current segment $E_{t:t+L}$ and those in the last segment $E_{t-L:t}$, which is the weight we assign to past motions $P_{t-L:t}$. The higher the similarity, the greater the weight we assign. This approach can not only encourage continuous transitions between two adjacent segments, but also ensure the maintenance of customized listener's behavioral habits between two adjacent sequences.

\vspace{-0.3cm}
\paragraph{Diffusion Model}
To achieve diverse and natural head motion generation, our generation module is built on denoising diffusion probabilistic models (DDPM) \cite{ddpm}.

During forward diffusion process $q$, we progressively inject noise into the listener motion $x_{0}$ until it becomes a Gaussian noise $x_{T}$, which can be defined as a Markov chain: 
\begin{equation}
    q(x_{1:T} \mid x_{0}) = \prod_{t=1}^{T} q(x_{t} \mid x_{t-1} ),
\end{equation}
\begin{equation}
    q(x_{t} \mid x_{t-1}) = \mathcal{N} (x_{t};\sqrt{1-\beta _{t}}x_{t-1}, \beta _{t}I),
\end{equation}
where $x_{1},...,x_{T}$ are noised listener motions, $\beta _{1},...,\beta _{T}$ are the variance schedule.
In the reverse process $p_{\theta }$, we produce listener motion $x_{0}$ by removing noise from $x_{T}$ step-by-step:
\begin{equation}
    p_{\theta } (x_{t-1} \mid x_{t},c) = \mathcal{N}(x_{t-1};\mu_{\theta} (x_{t},c,t), {\textstyle \sum_{\theta}^{}}(x_{t},c,t) ),
\end{equation}
where $c$ is a conditioning variable.  We perform denoising process with a transformer decoder \cite{transformer} conditioned on the motion prior and dynamic portrait tokens.

\vspace{-0.3cm}
\paragraph{Render} We utilize PIRenderer \cite{pirenderer} for rendering, which can produce natural listener videos by incorporating a single listener reference image as well as the 3DMM coefficients. 
\section{Experiments}
\label{sec:experiments}

\begin{table*}[t]
\footnotesize
\centering

\begin{tabular}{c cc cc cc cc cc cc cc}
\toprule
  \multirow{2}{*}{Methods} & \multirow{2}{*}{Testset} &\multicolumn{3}{c}{FD $\downarrow$} &   \multicolumn{2}{c}{RTLCC $\downarrow$} & 
  \multicolumn{2}{c}{RWTLCC$\downarrow$} &  
  \multicolumn{2}{c}{FID$_{\triangle fm}$$\downarrow$} & 
  \multicolumn{2}{c}{SND $\downarrow$} & \multicolumn{2}{c}{V-D $\uparrow$} \\

  \cmidrule(lr){3-5} \cmidrule(lr){6-7} \cmidrule(lr){8-9}
  \cmidrule(lr){10-11} \cmidrule(lr){12-13} \cmidrule(lr){14-15}

    & &  exp  &  angle &  trans &  exp  &  pose &  exp  &  pose &  exp  &  pose &  exp  &  pose &  exp  &  pose \\ \hline
    \multirow{2}{*}{RLHG$^{\ast}$ \cite{vico}} 
    &  $\mathcal{D}_{test}$ &  15.03 &  7.90 &  6.55 &  0.113 &  0.160 
    &  0.108 &  0.160 &  11.65 &  0.90 &  2.90 &  0.09  &  0.69 &  0.10  \\
    &  $\mathcal{D}_{ood}$ &  22.81 &  8.58 &  8.90 &  0.124 &  0.101 
    &  0.104 &  0.100 &  9.32 &  0.82 &  7.20 &  0.11  &  1.01 &  0.12\\
    \midrule
    
    \multirow{2}{*}{PCH$^{\ast}$ \cite{pchg}} 
    &  $\mathcal{D}_{test}$ &  19.02 &  13.26 &  7.74 &  0.124 &  0.163
    &  0.109 &  0.198 &  11.59 &  0.96 &  3.10 &  0.13  &  0.31 &  0.12 \\
    &  $\mathcal{D}_{ood}$ &  18.63 &  17.96 &  8.80 &  0.121 &  0.115 
    &  0.107 &  0.140 &  8.67 &  0.87 &  7.01 &  0.12 &  0.51 &  0.16 \\
    \midrule

    \multirow{2}{*}{L2L$^{\ast}$ \cite{l2l}} 
    &  $\mathcal{D}_{test}$ &  14.89 &  7.35 &  6.34 &  0.102 &  0.153 
    &  0.101 & 0.113  &  8.64 &  0.87 &  2.62 &  0.09  &  0.57 &  0.20 \\
    &  $\mathcal{D}_{ood}$ &  15.64 &  8.23 &  7.89 &  0.095 &  0.098 
    &  0.100 &  0.101 &  6.78 &  0.74 &  6.89 &  0.10  &  1.23 &  0.28 \\
    \midrule

    \multirow{2}{*}{MFR-Net$^{\dagger}$ \cite{mfr}} 
    &  $\mathcal{D}_{test}$ &  13.37 &  6.82 &  6.02 &  - &  - 
    &  - &  - &  - &  - &  - &  -  &  - &  - \\
    &  $\mathcal{D}_{ood}$ &  14.70 &  8.12 &  6.37 &  - &  - 
    &  - &  - &  - &  - &  - &  -  &  - &  - \\
    \midrule

    \multirow{2}{*}{Ours} 
    &  $\mathcal{D}_{test}$ &  \textbf{11.54} &  \textbf{6.12} &  \textbf{5.90} &  \textbf{0.072} &  \textbf{0.103} &  \textbf{0.081} &  \textbf{0.097} &  \textbf{3.12} &  \textbf{0.06} &  \textbf{2.40} &  \textbf{0.07} &  \textbf{1.55} &  \textbf{0.32}\\
    &  $\mathcal{D}_{ood}$ &  \textbf{12.67} &  \textbf{7.49} &  \textbf{6.01} &  \textbf{0.081} &  \textbf{0.073} 
    &  \textbf{0.084} &  \textbf{0.075} &  \textbf{3.59} & \textbf{0.07} &  \textbf{5.94} &  \textbf{0.05}  &  \textbf{1.25} &  \textbf{0.30} \\

\bottomrule
\end{tabular}
\caption{Comparisons of our model with other methods on $\mathcal{D}_{test}$ and $\mathcal{D}_{ood}$ of ViCo. \textbf{Bold} represents the best. The $^{\dagger}$ means we directly refer to data in their paper and $^{\ast }$ denotes we retrain the model. The $\downarrow$ indicates lower is better and the $\uparrow$ indicates higher is better. The values of FD and FID$_{\triangle fm}$ are multiplied by 100. The quantitative results on RealTalk \cite{realtalk} are in Appendix B.}
\label{table:main_table}
\end{table*}

\subsection{Datasets}
\paragraph{Data Annotation}
Since the existing listener head datasets do not contain the full data required for training our model, we conduct additional data annotations on two datasets:(1) ViCo \cite{vico}, a popular dataset in listener head generation; (2) RealTalk, proposed by \cite{realtalk}, which serves as a database for retrieving listener videos in \cite{realtalk}. Firstly, to learn the relationships between the text prior and listener expressions, we follow the text annotation pipeline in TalkCLIP \cite{talkclip}, and conduct text annotation on ViCo\cite{vico} and RealTalk\cite{realtalk}. Specifically, we employ \cite{libreface} to acquire emotion labels, activated AUs as well as their intensities. Additionally, to learn realistic head movements such as nodding to convey agreement and shaking the head to signify disagreement, we use Hopenet \cite{hopenet} to detect head motions. Finally, we generate diverse text prior description for each video segment according to the format: [A person $<$EMOTION$>$ and listens with $<$AU$>$ (and $<$HEAD MOTION$>$)], where $<$EMOTION$>$ denotes the emotional labels, $<$AU$>$ denotes fine-grained expression-related texts based on AU activation intensities and Facial Action Coding System \cite{FACS}, and $<$HEAD MOTION$>$ indicates head nodding/shaking labels. We use ``and" to link when there is more than one activated AU. Detailed examples can be seen in Appendix A.
\vspace{-0.5cm}
\paragraph{Data Construction}
For ViCo, which contains three parts: train set $\mathcal{D}_{train}$, test set $\mathcal{D}_{test}$, out-of-domain set $\mathcal{D}_{ood}$ (all identities in $\mathcal{D}_{ood}$ have not appeared in $\mathcal{D}_{train}$), we obtain the transcripts of the speaker's audios by a speech recognition model \cite{asr}, and get an extended version of ViCo. For RealTalk, to align the dimension of 3DMM coefficients with ViCo, we re-extract facial 3DMM coefficients using \cite{deep3d} and then obtain two subsets: $\mathcal{D}_{train}$ and $\mathcal{D}_{test}$. These two extended datasets are important for future research on text-guided listener head generation task. 

\begin{figure*}[t]
   \includegraphics[width=\linewidth]{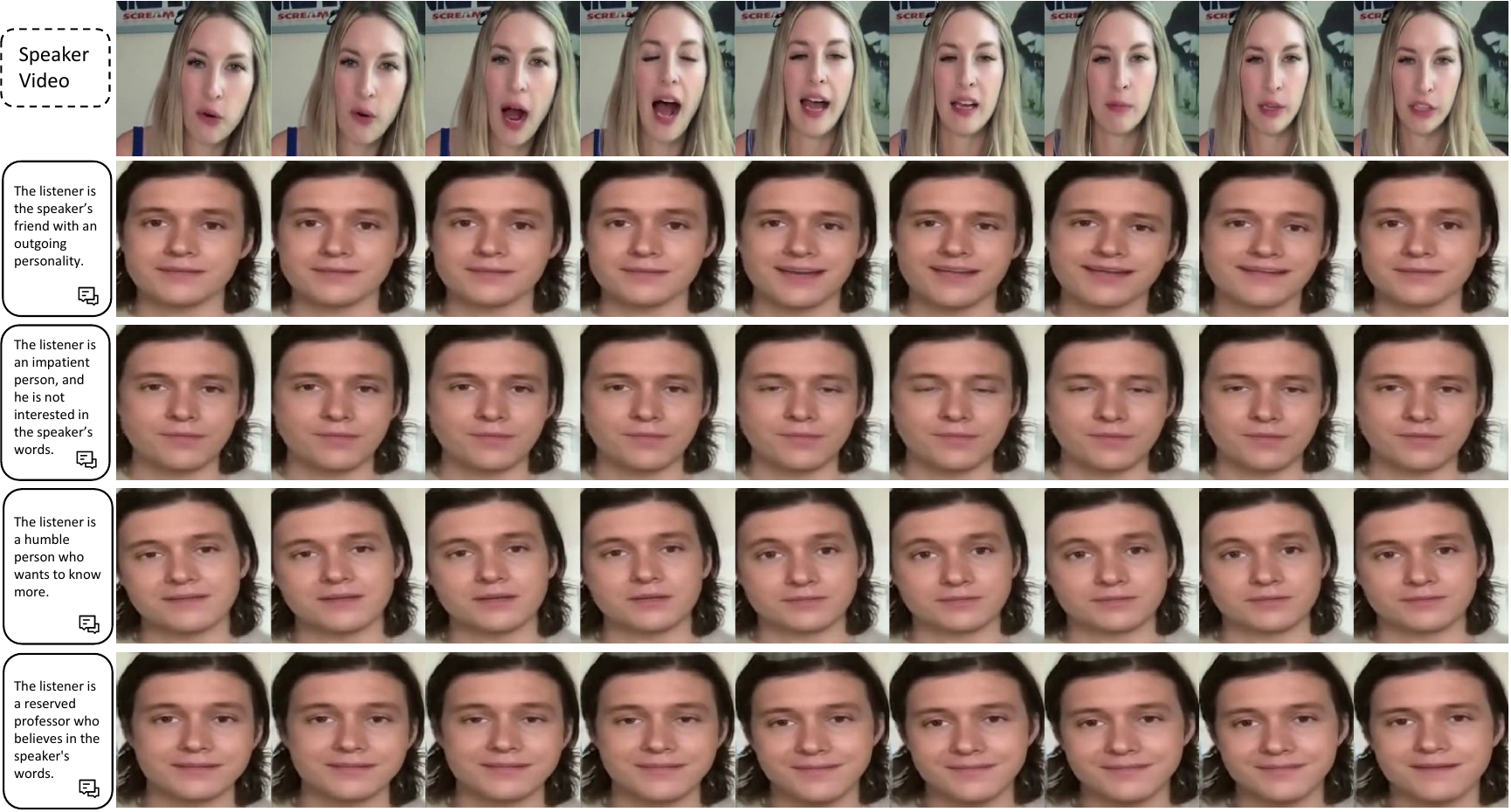}
   \caption{Visual results produced by CustomListener. All listener videos are generated conditioned on different pre-set text priors, the same speaker video (the 1st row) and the same reference listener image.}
   \vspace{-0.5cm}
   \label{fig:main_compare}
\end{figure*}

\subsection{Experimental Settings}
\paragraph{Loss Fuction} 
During training, followed by \cite{ddpm}, we perform denoising on noised listener motions $x_{t}$ conditioned on motion prior and dynamic portrait token step-by-step, and our optimization goal is to predict noise term $\epsilon$ at each step. The objective function can be defined as a conditional
diffusion loss as follows: 
\begin{equation}
\vspace{-0.1cm}
L_{d} = E_{x_{0},t,\epsilon _{t}\sim \mathcal{N}(0,I)} \left [ \left \| \epsilon _{t}-\epsilon _{\theta }(x_{t},c,t) \right \| ^{2}  \right ],
\vspace{-0.1cm}
\end{equation}
where $\epsilon _{t}\sim \mathcal{N}(0,\textbf{I})$ is the noise at step $t$, $c$ denotes conditions, $x_{0}$ is the original listener motions.

\vspace{-0.3cm}
\paragraph{Implementation Details} 
The resolution of videos is 256$\times$256, and the FPS is 30. For training, we clip the original videos to several 60-frame video segments. We extract 3DMM coefficients from speaker videos and listener videos by \cite{deep3d}, which serve as speaker motions and listener motions, respectively. Following MFR-Net \cite{mfr}, we extract 45-dim acoustic features from audio. We utilize AdamW optimizer \cite{adamw} to train our model with a learning rate of $1e^{-4}$.

\begin{table}[t]
\footnotesize
\centering
\begin{tabular}{c c c c c}
\toprule
    Method & SSIM $\uparrow$ & CPBD $\uparrow$ & PSNR $\uparrow$ & FID $\downarrow$ \\ \hline
    RLHG$^{\ast}$ \cite{vico} &  0.56  &  0.12  &  17.39  &  27.70 \\
    PCH$^{\ast}$ \cite{pchg} &  0.58  &  0.15  &  \textbf{18.48}  &  21.29 \\
    L2L$^{\ast}$ \cite{l2l} &  0.58  &  0.16  &  17.67 &  22.13 \\
    MFR-Net$^{\dagger}$ \cite{mfr} &  0.59  &  0.18  &  17.82  &  20.08 \\
    Ours &  \textbf{0.60}  &  \textbf{0.19}  &  17.98  &  \textbf{20.06} \\
      
\bottomrule
\end{tabular}
\caption{Quantitative results with state-of-the-art methods on image quality. The best results are highlighted in bold.}
\label{table:image_quality}
\end{table}

\subsection{Quantitative Results}
\paragraph{Metrics} It's non-trivial to measure the realism of generated listener motions. We choose the following metrics based on that a good custom listener should have four characteristics: (1) natural-looking and diverse facial/head motions; (2) highly synchronous with the speaker motions;  (3) highly coherent motions in a long video; (4) high image quality:

$\bullet$ FD: $L1$ distance between the predicted 70-dim listener motions and the ground-truth listener motions.

$\bullet$ Variation for Diversity (V-D) : 
Variance across the time series sequence of 3DMM coefficients.

$\bullet$ Residual Time Lagged Cross Correlation (RTLCC): $L1$ distance between generated TLCC \cite{TLCC_WTLCC} and TLCC \cite{TLCC_WTLCC} of ground truth. RTLCC indicates the correlation between listener motions and lagged speaker motions. 

$\bullet$ Residual Windowed Time Lagged Cross Correlation (RWTLCC): RTLCC in a fixed window. We set the window length to 4 seconds following ELP \cite{ELP}.

$\bullet$ FID$_{\triangle fm}$: FID score of differences in 3DMM coefficients between adjacent frames, which indicates the temporal naturalness of facial motions, followed by \cite{diffusion_priors}.

$\bullet$ SND: Sequence Naturalness Distance proposed by \cite{diffusion_priors}, showing the distribution difference between generated and ground-truth motions.

$\bullet$ SSIM, CPBD, PSNR and FID: Common metrics for image quality evaluation.

\vspace{-0.4cm}
\paragraph{Comparisons with Existing Methods} 
We retrain PCH \cite{pchg}, RLHG \cite{vico} and L2L \cite{l2l} on ViCo\cite{vico}. It should be noted that, source codes of ELP \cite{ELP} and MFR-Net \cite{mfr} is unavailable. For MFR-Net \cite{mfr}, we utilize the data from the original paper. For ELP \cite{ELP}, the 3DMM coefficient extraction model used is different from ours, resulting in distinct dimensions of coefficients (\textit{e.g.}, $\beta\in \mathbb{R}^{100T}$ for ELP \cite{ELP}, $\beta\in \mathbb{R}^{64T}$ for ours), thus it is not reasonable to directly compare with evaluation data in ELP \cite{ELP}. Therefore, we only provide its visual comparisons in Appendix C. 
Our quantitative comparisons consist of four parts:
1) precision and diversity of listener motions: as presented in Table \ref{table:main_table}, for the realism of listener motion generation, our proposed method achieves the lowest FD on both $\mathcal{D}_{test}$ and $\mathcal{D}_{ood}$ in ViCo. Meanwhile, the V-D of our model is comparable to other methods, which indicates that our model can maintain certain diversity when generating controllable listener motions. This may benefit from the probabilistic nature of the diffusion model. 
2) speaker-listener synchronization: compared to all existing methods, CustomListener can decrease the RTLCC and the RWTLCC, which justifies the effectiveness of our dynamic portrait tokens. 3) temporal smoothness: in Table \ref{table:main_table}, the FID$_{\triangle fm}$ and SND of our model outperform other existing methods by a large margin, which validates that our model can mitigate un-smooth transitions.
4) image-level quality: although image quality is not our main focus, we still have a certain performance gain in related metrics, as shown in Table \ref{table:image_quality}. These results indicate that our generated motion is closer to GT motion, thus achieving relatively high image-level quality without special design in rendering. 

\begin{figure}[t]
   \includegraphics[width=\linewidth]{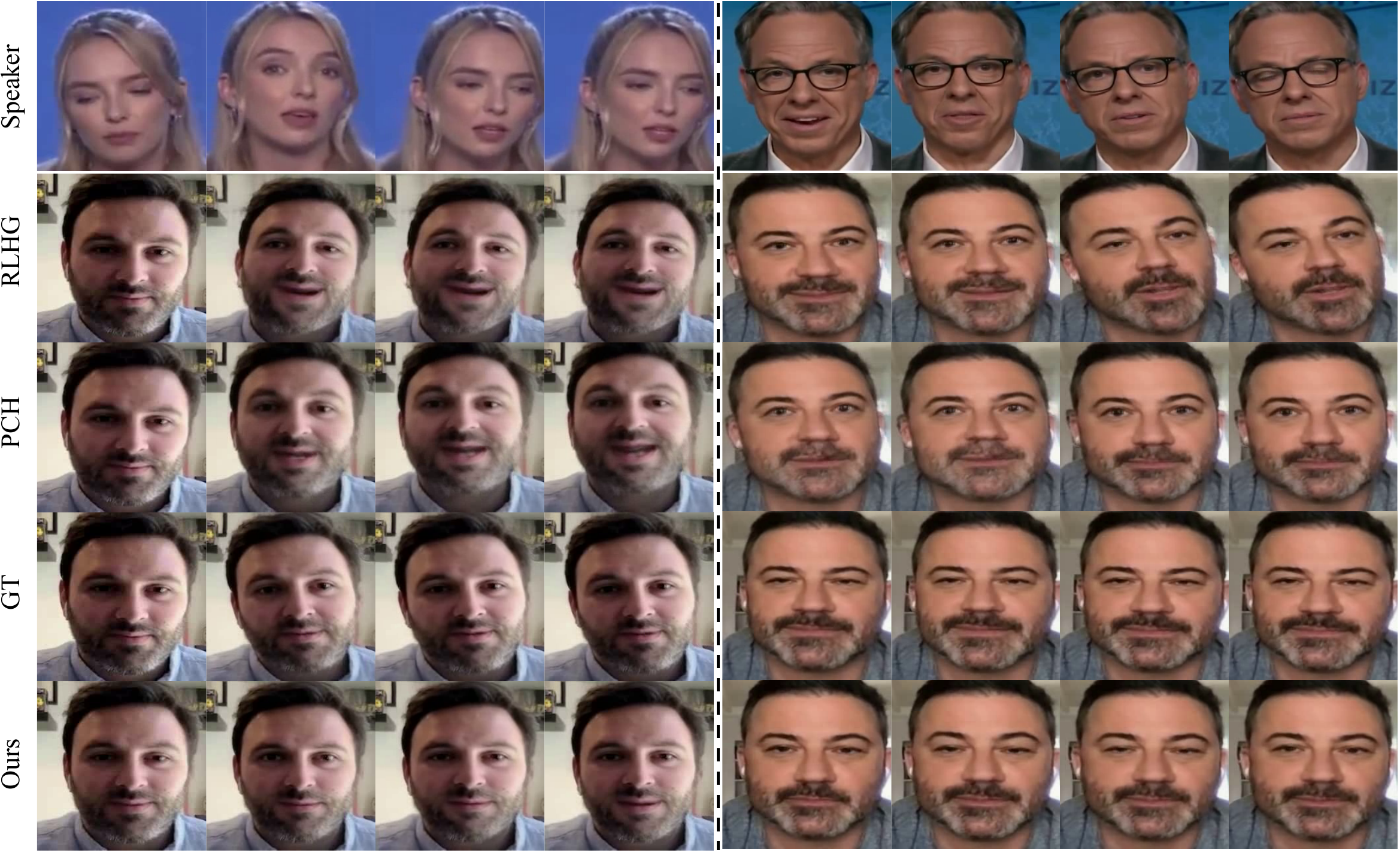}
   \caption{Qualitative comparisons with PCH \cite{pchg}, RLHG \cite{vico} conditioned on the same speaker and the same listener reference image.}
   \label{fig:visual_compare_with_others}
\end{figure}

\subsection{Qualitative Results}
\paragraph{Visual Results} 
Since previous methods cannot generate listener head based on fine-grained texts, we first present visual results generated by CustomListener. As shown in Figure \ref{fig:main_compare}, given different texts, covering diverse listener's personalities, identities and attitudes toward the speaker's utterances, our framework can produce fine-grained non-verbal listener motions closely aligned with pre-customized texts. For example, if the listener is an impatient person who is not interested in the speaker’s words(the 3rd row in Figure \ref{fig:main_compare}), he may appear indifferent and show impatience. Additionally, the produced listener’s responses can fluctuate with the speaker's semantics and movement, showing great effectiveness of our proposed dynamic portrait tokens.

\begin{figure}[t]
   \includegraphics[width=\linewidth]{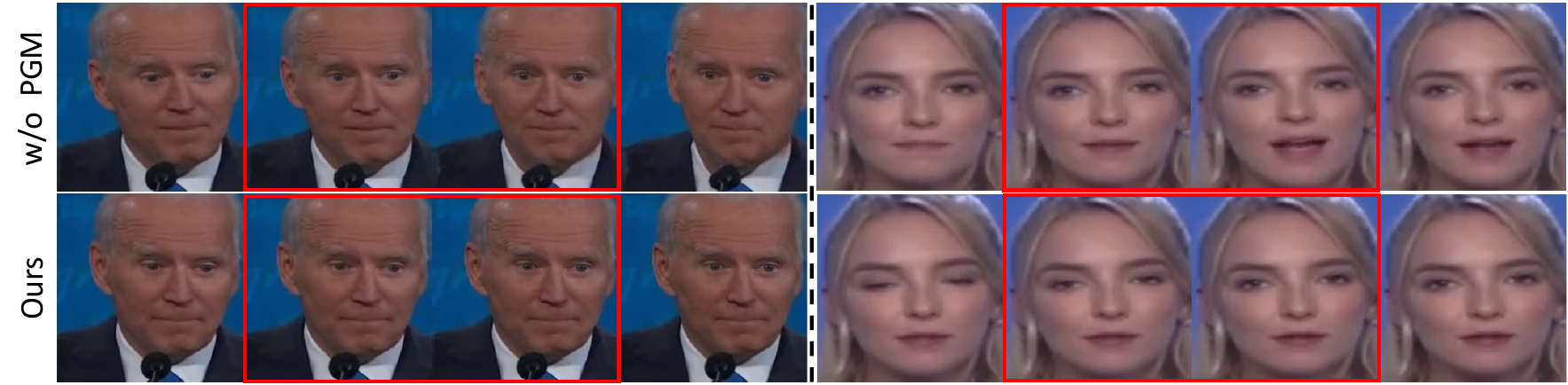}
   \caption{Ablation study of PGM to verify that PGM can maintain motion coherency. Each group includes adjacent four frames, and the frames enclosed by the red box are the two frames at the transition point between two video clips.}
   \vspace{-0.5cm}
   \label{fig:PGM_ablation}
\end{figure}
\begin{table}[t]
\footnotesize
\centering
\setlength\tabcolsep{0.15cm}
\begin{tabular}{c c c c c c}
\toprule
    \multirow{2}{*}{Method} & FD $\downarrow$ & RTLCC $\downarrow$ & 
    RWTLCC $\downarrow$ &  FID $_{\triangle fm}$$\downarrow$ & 
    SND $\downarrow$ \\ 

    & (x100)  &   &   &  (x100)  &  \\
    
    \hline
    w/o ATRI &  20.67 &  0.206 &  0.203 &  5.01  &  4.53 \\
    w/o SMW &  19.05 &  0.201 &  0.196  &  5.07  &  4.58 \\
    w/o SDP &  21.97 &  0.213 &  0.210 &  5.90  &  4.60 \\
    w/o PGM &  20.53 &  0.197 &  0.201 &  6.47  &  4.65 \\
    Ours &  18.48 &  0.165 &  0.169 &  3.42  &  4.23 \\

\bottomrule
\end{tabular}
\caption{Ablation study. Each cell represents the average results of 70-dim motions (64-dim exp + 6-dim pose) on $\mathcal{D}_{test}$ and $\mathcal{D}_{ood}$.}
\label{table:ablation}
\end{table}

\vspace{-0.4cm}
\paragraph{Comparisons with Existing Methods} 
In Figure \ref{fig:visual_compare_with_others}, we compare our results with RLHG \cite{vico}, PCH \cite{pchg}. It can be seen that our method is capable of generating more realistic listener heads than other baselines. In the left four columns, both RLHG \cite{vico} and PCH \cite{pchg} suffer from unnatural motions (\textit{e.g.}, disorderly opened mouth) and visible facial artifacts. In the right four columns, RLHG \cite{vico} still exhibits incorrect motions and PCH \cite{pchg} can not display negative attitude accurately. While our method can produce natural listener head that highly aligns with the ground truth conditioned on the given texts. More detailed comparisons can be seen in Appendix C.

\vspace{-0.4cm}
\paragraph{User Study} 
We conduct user studies to evaluate from human perceptual perspective. We randomly choose 25 speaker videos from ViCo\cite{vico} and RealTalk\cite{realtalk} to generate listeners for each
method. 28 people are asked to rate different methods(1-4, 4 is the best) for each video from 4 metrics: similarity to GT, overall naturalness, speaker-listener coordination and video smooth. The average scores of each method are shown in Table \ref{table:user_study}. Our method outperformed other methods in all aspects. 
\vspace{-0.4cm}
\begin{table}[t]
\footnotesize
\centering
\begin{tabular}{c c c c c}
\toprule
    \multirow{2}{*}{Method}  & Similarity  & Overall & Speaker-listener  & Video \\ 
    & to GT & Naturalness & Coordination & Smooth \\
    \hline
    RLHG \cite{vico} &  1.28  &  1.12  &  1.64  &  2.04 \\
    PCH \cite{pchg} &  2.08  &  2.40  &  2.48  &  2.28 \\
    L2L \cite{l2l} &  2.68  &  2.56  &  1.92  &  1.80 \\
    Ours &  \textbf{3.96}  &  \textbf{3.92}  &  \textbf{3.96}  &  \textbf{3.88} \\
  
\bottomrule
\end{tabular}
\caption{User study results. The best results are highlighted in bold.}
\label{table:user_study}
\end{table}

\begin{figure}[t]
   \includegraphics[width=\linewidth]{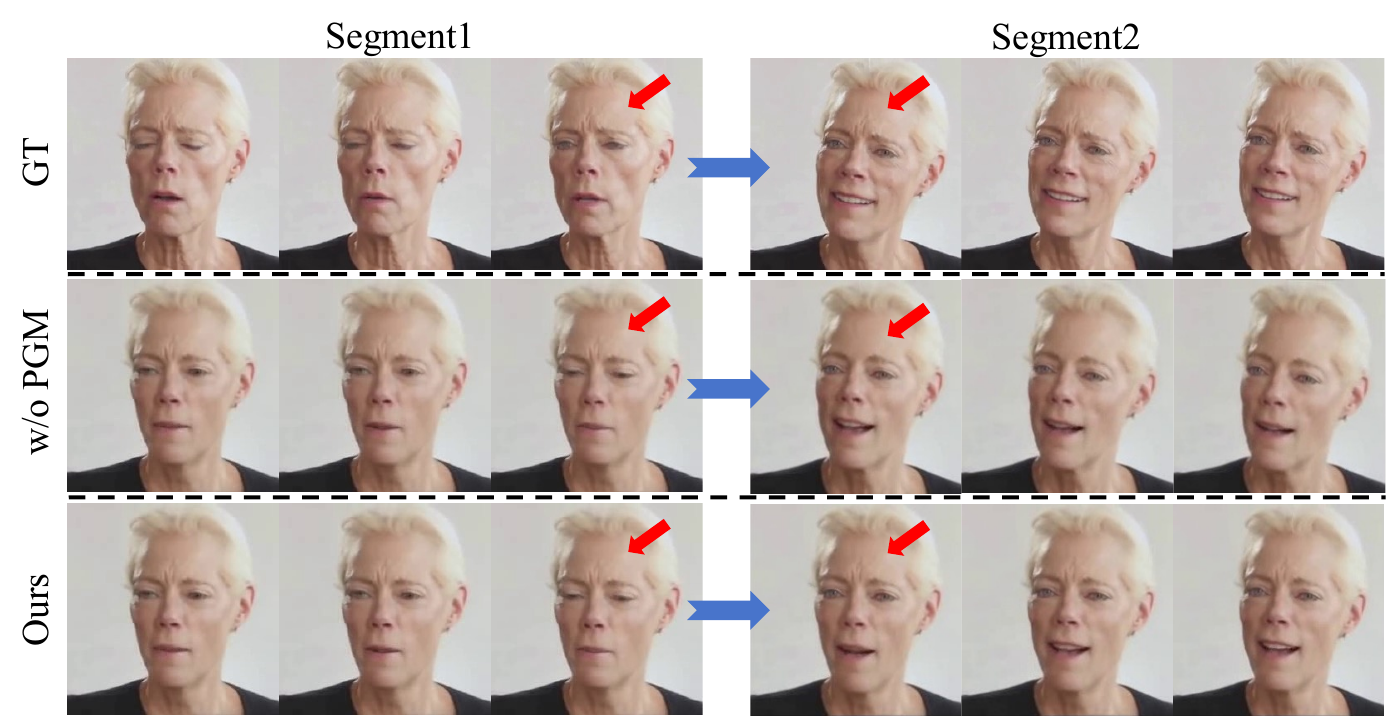}
   \caption{Ablation study of PGM. For different segments, the habit of frowning is maintained using PGM.}
   \vspace{-0.5cm}
   \label{fig:PGM_visual}
\end{figure}

\subsection{Ablation Study}
\label{sec:ablation}
\paragraph{SDP module}
In Section \ref{subsec:3.2}, we design a SDP module to generate dynamic portrait tokens for realizing progressive motion shifts that correspond well to the speaker. To validate its effectiveness, we conduct ablation study on Audio-text Responsive Interaction (ATRI) and Speaker Motion-based Weighting (SMW). The results are shown in Table \ref{table:ablation}. In addition to the improvement in FD, the improvement in the RTLCC and RWTLCC is more pronounced, which demonstrates the SDP module is crucial for enhancing speaker-listener synchronization, and each of its components matters.

\vspace{-0.4cm}
\paragraph{Past-guided module} To validate the effectiveness of our Past-guided module (PGM), we train networks without PGM. As presented in Table \ref{table:ablation}, without PGM, the metrics FID$_{\triangle fm}$ and SND, which are related to the temporal naturalness of motions, increase significantly, indicating a decrease in smoothness. This observation shows the necessity of PGM for mitigating unnatural transitions between video clips. The visual results are shown in Figure \ref{fig:PGM_ablation}. Apart from this, our proposed PGM can maintain the consistency of listener’s customized behavioral habits between adjacent clips, as shown in Figure \ref{fig:PGM_visual}. The listener is customized a habit of frowning when thinking, and when segment 2 is guided by ``lips mildly parted and lip corner raised slightly", the frowning habit can be still maintained.

\vspace{-0.25cm}
\section{Conclusion}
\vspace{-0.15cm}
In this paper, we present a user-friendly framework called CustomListener, which can 
realize freely-controllable listener head generation conditioned on text priors. To achieve the speaker-listener coordination, we propose a SDP-module for dynamic portrait-token generation, which is capable of representing progressive motion changes. To ensure coherent listener responses in a long video, we design a PGG-module for smooth transition between adjacent video segments as well as maintaining customized behavioral habits. Comprehensive experiments demonstrate the superiority of our whole framework. In future work, it may be interesting to generalize our framework to listening body generation, yielding a complete and realistic listener agent.

{
    \small
    \bibliographystyle{ieeenat_fullname}
    \bibliography{main}
}
\clearpage
\maketitlesupplementary
\appendix

\section{Implementation Details}
\subsection{Network Details}
\begin{figure}[t]
   \includegraphics[width=0.48\textwidth]{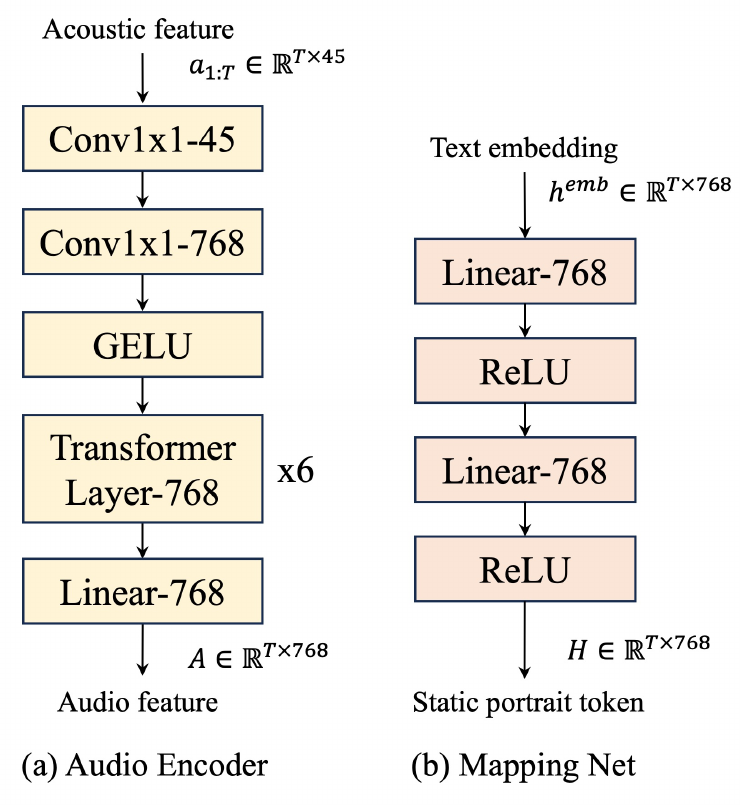}
   \caption{Stucture of Audio Encoder and Mapping Net.}
   \vspace{-0.5cm}
   \label{fig:supp_net_sturc}
\end{figure}

We introduced Audio Encoder, Mapping Net, Motion Encoder and Transformer Decoder-based Diffusion Model in Section \ref{sec:method} of our main paper. Due to the page limitation, we show implementation details of these modules in this Appendix.
\vspace{-0.3cm}
\paragraph{Audio Encoder}
For audio preprocessing, we first extract 45-dim acoustic features from speaker audio, and then encode these acoustic features into audio features. The structure of Audio Encoder is shown in Figure \ref{fig:supp_net_sturc}, which consists of two convolution layers, a GELU activation function, six 768-dim transformer layers, and a linear layer.

\vspace{-0.3cm}
\paragraph{Mapping Net}
As discussed in Section \ref{subsec:3.2} of the main paper, in order to obtain portrait-related tokens, we further employ a Mapping Net to convert text embeddings from RoBERTa \cite{roberta} into static portrait tokens, which  describes the listener’s static portrait (e.g. expressions, head movements) in the entire video segment. The structure of Mapping Net is shown in Figure \ref{fig:supp_net_sturc}.

\vspace{-0.3cm}
\paragraph{Motion Encoder} We only use a single linear layer to encode the 70-dim noised listener motions to 768-dim motion tokens (denoted with small blue squares in Figure \ref{fig:main} of our main paper).

\vspace{-0.3cm}
\paragraph{Transformer Decoder-based Diffusion Model} We employ transformer decoder as our basic denoising module for two reasons: First, compared to U-Net \cite{unet} used in traditional diffusion model, transformer \cite{transformer} can better capture temporal information in listener motion sequences. Furthermore, compared to transformer encoder which only contains self-attention, transformer decoder is capable of facilitating information interaction better between noised listener motion features as well as concatenated dynamic portrait token and motion prior. Specifically, we use an 8-layer transformer decoder, with 8 heads and 2048-dim fully forward layers. Additionally, we concatenate the input motion tokens with diffusion time-step token (denoted with small pink squares in Figure \ref{fig:main} of our main paper) to inject time-step information.

\subsection{Inference Details}
\paragraph{Long-video Inference}
During inference, we generate long-term video in a segment-by-segment way. Specifically, we first sequentially clip the original long video to several 60-frame video segments, containing video frames, audios and speech content. We then input user-customized listener attributes. Subsequently, by incorporating speech content of each segment as well as pre-customized attribute description, text priors of each segment will be produced with the help of GPT. Finally, our proposed CustomListener will generate long-term listener motions clip-by-clip.

\begin{figure*}[t]
   \includegraphics[width=\linewidth]{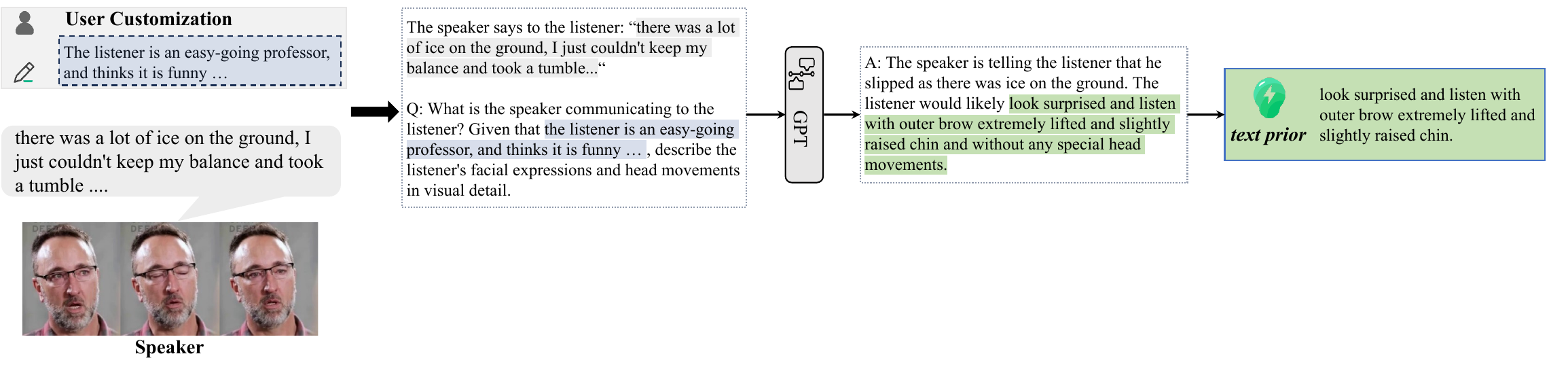}
   \caption{Details of Text Prior Generation.}
   \label{fig:text_prior_gen}
\end{figure*}

\begin{figure*}[t]
   \includegraphics[width=\linewidth]{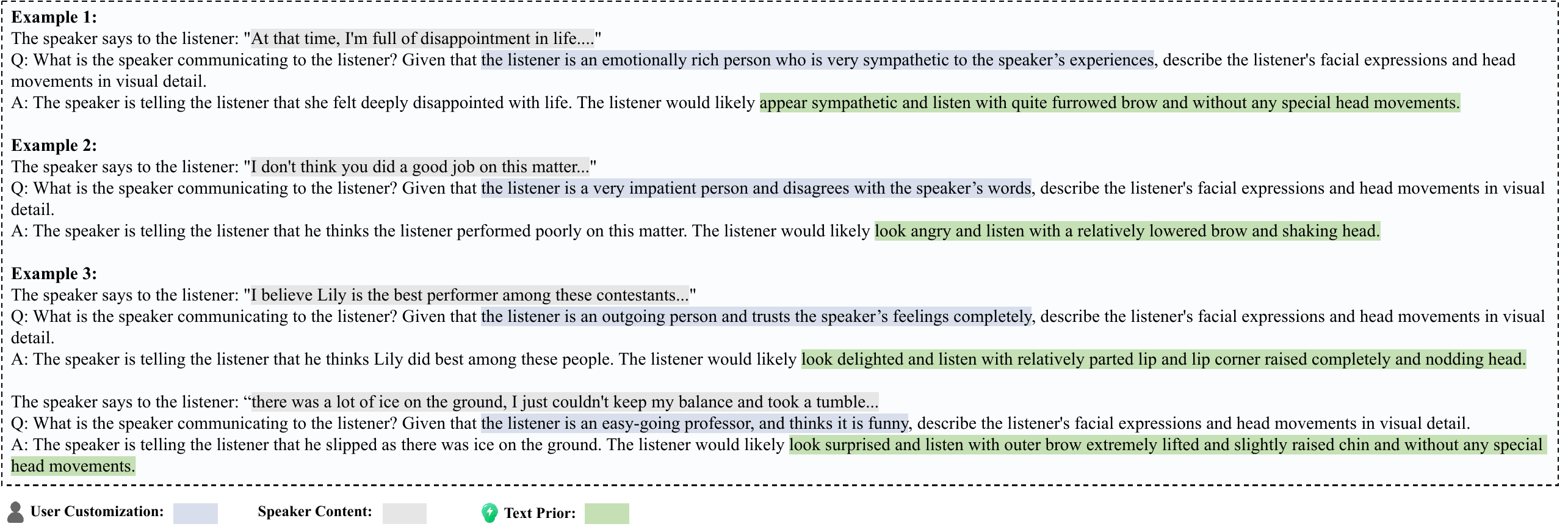}
   \caption{Illustration of Chain-of-thought 
Prompting.}
   \vspace{-0.3cm}
   \label{fig:gpt_prompt_example}
\end{figure*}

\paragraph{Text Prior Generation}
To generate text prior in specific format like [A person $<$EMOTION$>$ and listens with $<$AU$>$ (and $<$HEAD MOTION$>$)]), inspired by \cite{realtalk}, we use chain-of-thought prompting \cite{chain_of_thought} when querying GPT. Specifically, before asking question, we first provide GPT with a few examples showing the desired answer format as well as reasoning process. This strategy can also enable GPT to provide more accurate text prior about facial expressions and head motions while reducing the misunderstanding probability towards the speaker's content. We present some examples of queries in Figure \ref{fig:text_prior_gen} and Figure \ref{fig:gpt_prompt_example}.

After getting the text prior, we randomly replace the adjectives and adverbs to enhance the diversity of description. For example, for $<$EMOTION$>$, if it is \textquotesingle happy\textquotesingle, we replace it by randomly selecting one from a group $<$looks happy, seems joyful, ...$>$. For $<$AU$>$, following Facial Action Coding System, if AU12 (lip corner puller) is activated and level=3, we randomly choose an adj. from $<$raised, pulled, ...$>$, an adv. from $<$fully, extremely, ...$>$, then combine the above and get [ A person seems joyful and listens with fully raised lip corners.].
\begin{table*}[h]
\footnotesize
\centering
\vspace{2mm}

\begin{tabular}{c c cc cc cc cc cc cc}
\toprule
  \multirow{2}{*}{Methods} &\multicolumn{3}{c}{FD $\downarrow$} &   \multicolumn{2}{c}{RTLCC $\downarrow$} & 
  \multicolumn{2}{c}{RWTLCC$\downarrow$} &  
  \multicolumn{2}{c}{FID$_{\triangle fm}$$\downarrow$} & 
  \multicolumn{2}{c}{SND $\downarrow$} & \multicolumn{2}{c}{V-D $\uparrow$} \\

  \cmidrule(lr){2-4} \cmidrule(lr){5-6} \cmidrule(lr){7-8}
  \cmidrule(lr){9-10} \cmidrule(lr){11-12} \cmidrule(lr){13-14}

    & exp  &  angle &  trans &  exp  &  pose &  exp  &  pose &  exp  &  pose &  exp  &  pose &  exp  &  pose \\ \hline
    RLHG$^{\ast}$ \cite{vico}
    &  20.11 &  10.32 &  7.16 &  0.162 &  0.365
    &  0.160 &  0.363 &  12.42 &  0.76 &  4.75 &  0.09  &  0.54 &  0.14  \\
    \midrule
    
    PCH$^{\ast}$ \cite{pchg}
    &  23.07 &  13.46 &  8.04 &  0.170 &  0.368
    &  0.163 &  0.391 &  12.38 &  0.81 &  5.02 &  0.11  &  0.29 &  0.10 \\
    \midrule

    L2L$^{\ast}$ \cite{l2l}
    &  19.01 &  10.16 &  7.03 &  0.158 &  0.319 
    &  0.156 & 0.315  &  10.80 &  0.71 &  4.67 &  0.10  &  0.56 &  0.40 \\
    \midrule

    Ours
    &  \textbf{17.63} &  \textbf{9.30} &  \textbf{6.49} &  \textbf{0.138} &  \textbf{0.218} &  \textbf{0.134} &  \textbf{0.207} &  \textbf{7.70} &  \textbf{0.14} &  \textbf{4.37} &  \textbf{0.07} &  \textbf{2.90} &  \textbf{1.01}\\

\bottomrule
\end{tabular}
\caption{Comparisons of our model with other methods on $\mathcal{D}_{test}$ of RealTalk. \textbf{Bold} represents the best. The $^{\ast }$ denotes we retrain the model. The $\downarrow$ indicates lower is better and the $\uparrow$ indicates higher is better. The values of FD and FID$_{\triangle fm}$ are multiplied by 100.}
\label{table:realtalk_main}
\end{table*}
\begin{table}[h]
\footnotesize
\centering
\vspace{2mm}
\begin{tabular}{c c c c c}
\toprule
    Method & SSIM $\uparrow$ & CPBD $\uparrow$ & PSNR $\uparrow$ & FID $\downarrow$ \\ \hline
    RLHG$^{\ast }$ \cite{vico} &  0.51  &  0.29  &  15.34  &  30.22 \\
    PCH$^{\ast }$ \cite{pchg} &  0.56  &  0.32  &  \textbf{16.13}  &  24.68 \\
    L2L$^{\ast }$ \cite{l2l} &  0.56  &  0.31  &  15.98  &  25.68 \\
    Ours &  \textbf{0.58}  &  \textbf{0.34}  &  16.01  &  \textbf{23.86} \\
\bottomrule
\end{tabular}
\caption{Quantitative results with other methods on image quality. The $^{\ast }$ denotes we retrain the model. The best results are highlighted in bold. }
\label{table:realtalk_image_quality}
\end{table}
\section{Quantitative Results on RealTalk}
We did not present quantitative results on RealTalk in main paper due to page limitation, thus we give these results in the Appendix B. For a fair comparison, we retrain PCH \cite{pchg}, RLHG \cite{vico} and L2L \cite{l2l} on RealTalk dataset. Due to the unavailability of source codes, we cannot compare with MFR-Net \cite{mfr} and ELP \cite{ELP} on RealTalk. Notably, since PCH \cite{pchg} and RLHG \cite{vico} need listener’s attitude label as input, we use \cite{libreface} to obtain emotion labels of each video clip. Then, we group these emotion labels into three basic attitudes: positive, negative and neutral. As shown in Table \ref{table:realtalk_main}, we present the FD, RTLCC, RWTLCC, FID$_{\triangle fm}$, SND and V-D of our model as well as other methods. Our proposed CustomListener achieves best performance on RealTalk across all metrics, which demonstrates the superiority of CustomListener in generating realistic and coherent listener motions that highly synchronous with the speaker's motions. In Table \ref{table:realtalk_image_quality}, we also present evaluation results related to image-level quality on RealTalk. Our model achieves optimal results, which indicates CustomListener can generate relatively high-quality listener images without special design in the render module, thus justifies our generated motions are highly precise.
\section{Supplementary Visual Results}
\subsection{Visual Comparisons with Other Methods}
\begin{figure}[t]
   \includegraphics[width=\linewidth]{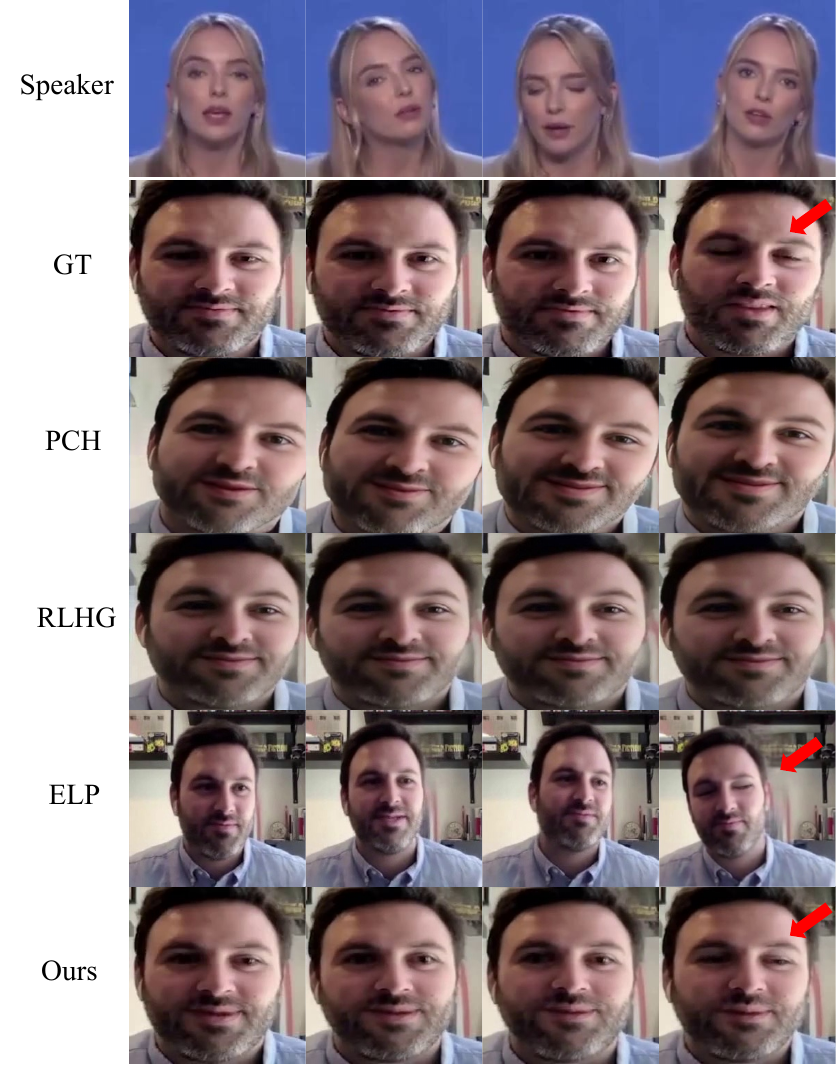}
   \caption{Qualitative comparisons with PCH \cite{pchg}, RLHG \cite{vico} and ELP \cite{ELP}.}
   \vspace{-0.5cm}
   \label{fig:supp_ELP_compare}
\end{figure}

\begin{figure}[t]
   \includegraphics[width=\linewidth]{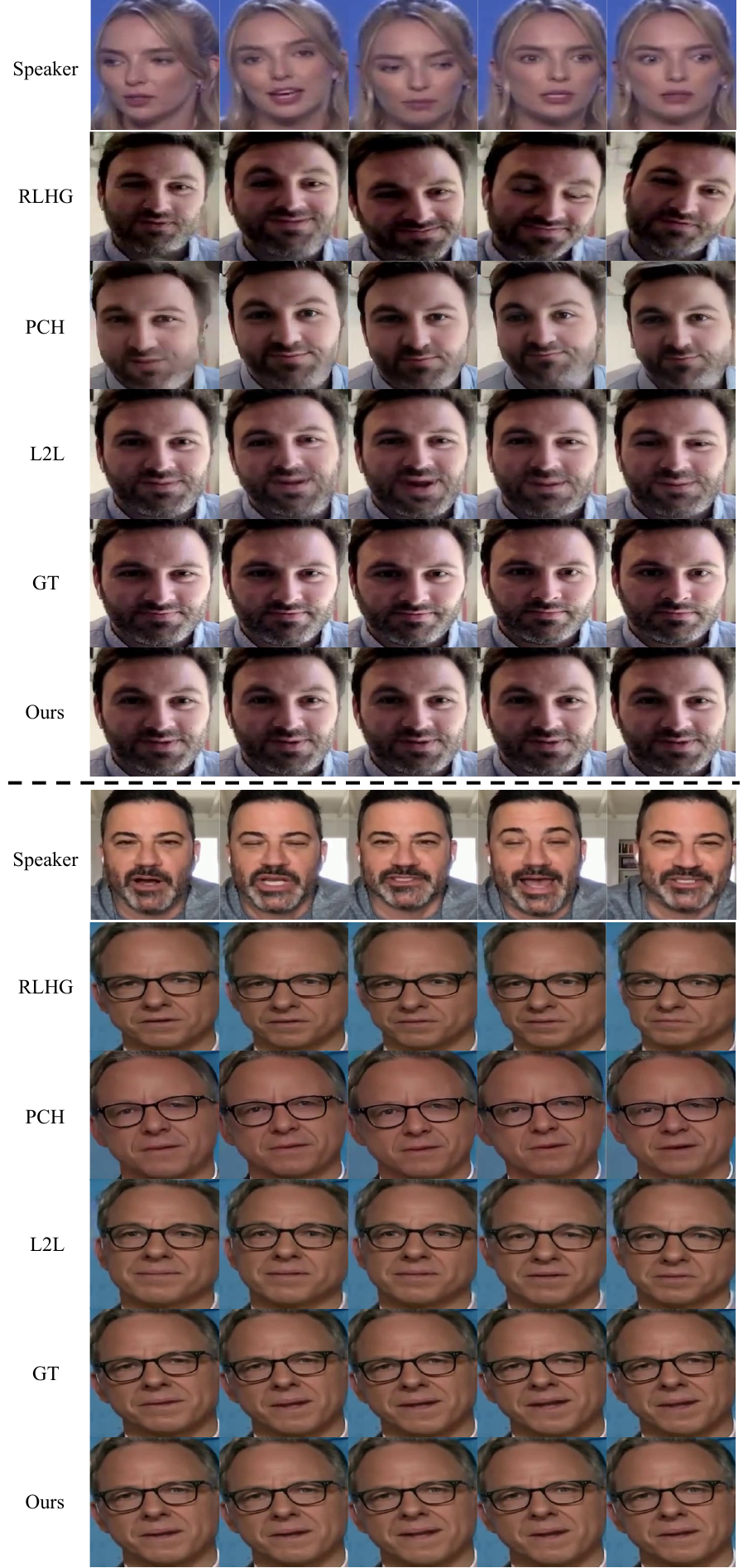}
   \caption{Qualitative comparisons with other methods on ViCo dataset.}
   \vspace{-0.5cm}
   \label{fig:supp_vico_compare}
\end{figure}

\begin{figure}[t]
   \includegraphics[width=\linewidth]{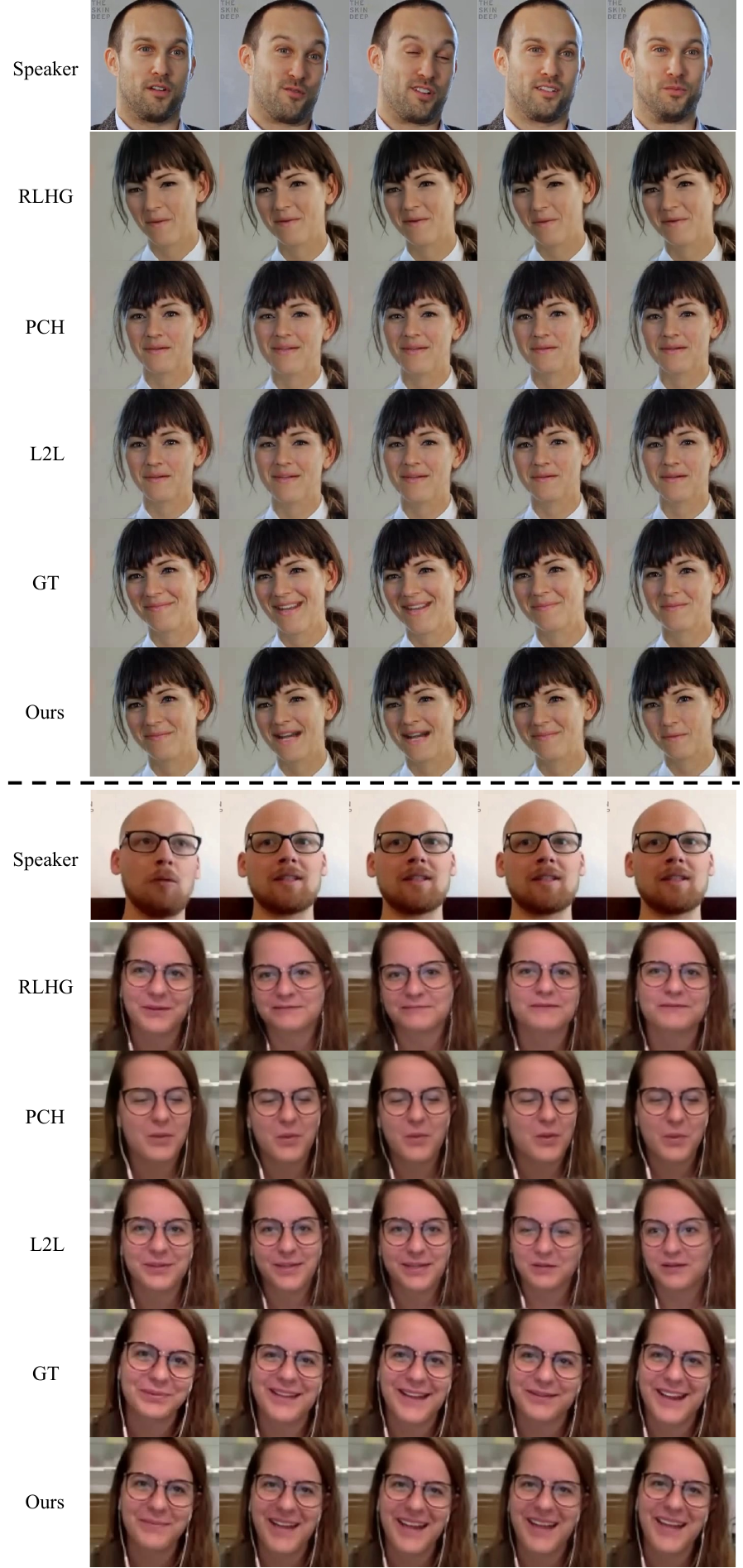}
   \caption{Qualitative comparisons with other methods on RealTalk dataset.}
   \vspace{-1cm}
   \label{fig:supp_realtalk_compare}
\end{figure}

In Figure \ref{fig:supp_ELP_compare}, we compared our results with PCH \cite{pchg}, RLHG \cite{vico} and ELP \cite{ELP}. 
As there is no source code provided by ELP \cite{ELP}, for fairly comparison, we first find the speaker video showed in ELP \cite{ELP}, and then generate listener head based on the speaker video and corresponding description texts instead of an attitude label. Then, we displayed four generated listener frames, which are aligned with the four speaker frames in time. As shown in the visual comparisons, our method has two advantages. Firstly, without any special design in render model, our generated listener images showed less facial artifacts and more closed to the ground truth, which indicates the listener motions produced by CustomListener are highly aligned with the ground-truth motions. Secondly, without a blink modeling like ELP \cite{ELP}, our generated listener can perform blinking action, which demonstrates that the listener motions are natural-looking and photorealistic. More visual comparisons with 
PCH \cite{pchg}, RLHG \cite{vico} and L2L \cite{l2l} on ViCo and RealTalk are presented in Figure \ref{fig:supp_vico_compare} and Figure \ref{fig:supp_realtalk_compare}, respectively.

\begin{figure*}[t]
   \includegraphics[width=\linewidth]{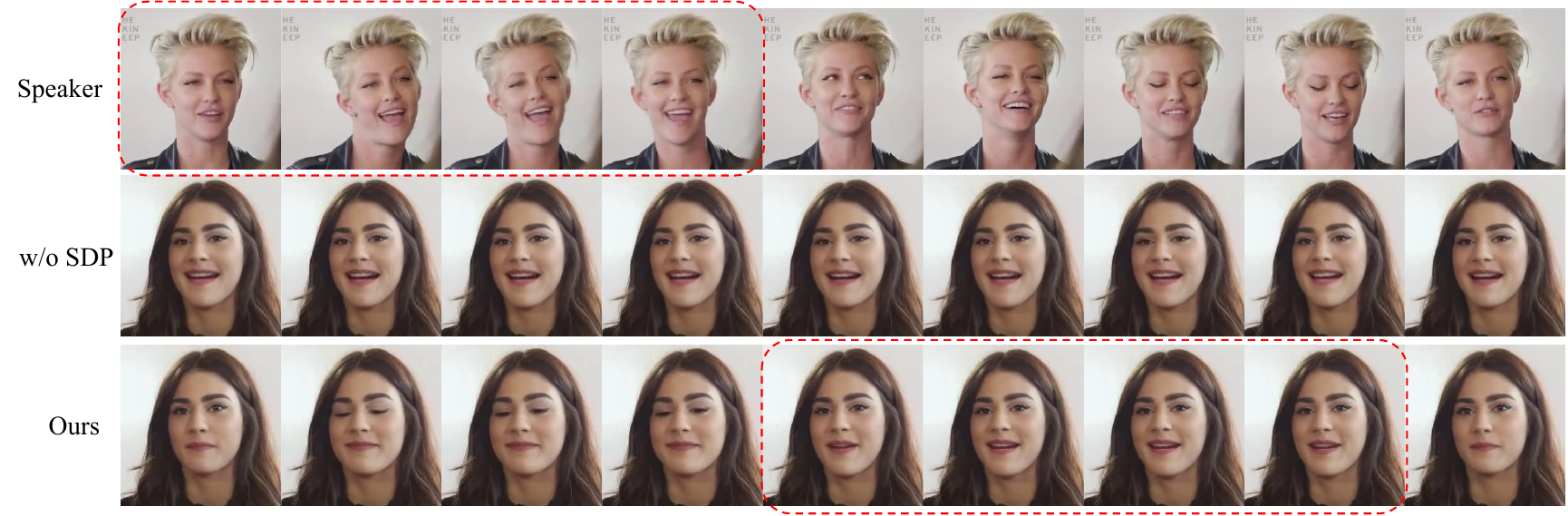}
   \caption{Ablation study of SDP module.}
   \label{fig:ablation_SDP}
\end{figure*}

\subsection{Visual Results of Ablation}
In Section \ref{sec:ablation} of our main paper, we presented visual ablation of PGM module. Due to the page limitation, we show the visual ablation of SDP module in this subsection. As shown in Figure \ref{fig:ablation_SDP}, with the help of SDP module, our generated listener motions can not only realize progressive motion changes, but also fluctuate with the speaker’s semantics, intonation and movement amplitude, and thus achieve the speaker-listener coordination.

\subsection{Supplementary Video}
We provide a supplementary video to present more visual results, which include visual videos on ViCo and RealTalk, more visual comparisons with other methods as well as ablation study of each component.  
The video can be found in the ``Supplementary\_Videos.mp4" in the Supplementary Materials. These videos are also displayed on the project page.

\end{document}